\documentclass[journal]{IEEEtran}

\ifCLASSINFOpdf
  
\else

\fi

\ifCLASSOPTIONcompsoc
\usepackage[nocompress]{cite}
\else
\usepackage{cite}
\fi
\usepackage{amsmath}
\usepackage{algorithmic}
\usepackage{algorithm}
\usepackage{array}
\usepackage{subfigure}
\usepackage{multirow}
\usepackage{booktabs}
\usepackage{threeparttable}
\usepackage{stfloats}
\usepackage{amsfonts,amssymb}
\usepackage{color}
\usepackage{bbm}
\usepackage{bm}
\usepackage{ragged2e}
\usepackage{url}
\usepackage{graphicx}

\begin{document}
%
\title{Causal Mechanism Transfer Network for Time Series Domain Adaptation in Mechanical Systems}
%
%
%


\author{Zijian Li,
	Ruichu Cai*,~\IEEEmembership{Member,~IEEE,}
	Kok Soon Chai,
	Hong Wei Ng, 
	Hoang Dung Vu, 
	Marianne Winslett,
	Tom Z. J. Fu,
	Boyan Xu,
	Xiaoyan Yang,
	Zhenjie Zhang
\thanks{Manuscript received XX; revised XX; accepted XX. Date of publication XX XX, 2019; date of current version XX XX, 2019. Ruichu and Zijian were supported in part by NSFC-Guangdong Joint Found (U1501254), Natural Science Foundation of China (61876043), Natural Science Foundation of Guangdong (2014A030306004, 2014A030308008), Guangdong High-level Personnel of Special Support Program (2015TQ01X140), Science and Technology Planning Project of Guangzhou (201902010058).
Tom Fu is partially supported by Natural Science Foundation of China (61702113) and China Postdoctoral Science Foundation (2017M612613).
And it is also supported by the National Research Foundation, Prime Minister's Office, Singapore under its Campus for Research Excellence and Technological Enterprise (CREATE) programme. It is also supported by the Republic of Singapore's National Research Foundation (NRF) through Building and Construction Authority (BCA)'s Green Buildings Innovation Cluster (GBIC) R\&D Grant, BCA RID 94.17.2.8. It is also partially supported by SK Telecom.  (\emph{Corresponding author: Ruichu Cai.})
}
\thanks{Zijian Li, Ruichu Cai, Boyan Xu are with the School of Computer, Guangdong University of Technology, Guangzhou, China, 510006. \protect E-mail: \{leizigin, cairuichu, hpakyim\}@gmail.com.}
\thanks{Kok Soon Chai and Hoang Dung Vu are with Kaer Pte. Ltd, Singapore. \protect E-mail: \{koksoon.chai, jose.vu\}@kaer.com.}
\thanks{Marianne Winslett, Tom Z. J. Fu and Hong Wei Ng are with Advanced Digital Science Center, Singapore. \protect E-mail: winslett@illinois.edu, Tom.fu@adsc.com.sg, lightalchemist@gmail.com.}
\thanks{Xiaoyan Yang, Zhenjie Zhang are with Yitu Technology, Singapore R\&D. \protect E-mail:\{xiaoyan.yang, zhenjie.zhang\}@yitu-inc.com}

}

\markboth{IEEE TRANSACTIONS ON CYBERNETICS,~Vol.~14, No.~8, JUNE~2019}%
{Shell \MakeLowercase{\textit{et al.}}: Bare Demo of IEEEtran.cls for IEEE Journals}
\maketitle

\begin{abstract}
\justifying Data-driven models are becoming essential parts in modern mechanical systems, commonly used to capture the behavior of various equipment and varying environmental characteristics. 
Despite the advantages of these data-driven models on excellent adaptivity to high dynamics and aging equipment, they are usually hungry to massive labels over historical data, mostly contributed by human engineers at an extremely high cost. The label demand is now the major limiting factor to modeling accuracy, hindering the fulfillment of visions for applications. 
Fortunately, domain adaptation enhances the model generalization by utilizing the labelled source data as well as the unlabelled target data and then we can reuse the model on different domains. However, the mainstream domain adaptation methods cannot achieve ideal performance on time series data, because most of them focus on static samples and even the existing time series domain adaptation methods ignore the properties of time series data, such as temporal causal mechanism. 
In this paper, we assume that causal mechanism is invariant and present our Causal Mechanism Transfer Network(\textbf{CMTN}) for time series domain adaptation. 
By capturing and transferring the dynamic and temporal causal mechanism of multivariate time series data and alleviating the time lags and different value ranges among different machines, \textbf{CMTN} allows the data-driven models to exploit existing data and labels from similar systems, such that the resulting model on a new system is highly reliable even with very limited data. 
We report our empirical results and lessons learned from two real-world case studies, on chiller plant energy optimization and boiler fault detection, which outperforms the existing state-of-the-art method. 
\end{abstract}


%
\IEEEpeerreviewmaketitle

\section{Introduction}
\IEEEPARstart {E}{xplosively} growing data from Internet of Things (IoT) are now flooding into data management systems for processing and analysis. The availability of massive historical data, powerful deep learning frameworks and excessive computation power are together boosting the development of new data-driven models over complex mechanical systems, which are used to characterize the behaviors of the systems running with highly dynamic external conditions and aging equipment.

While data-driven models have achieved significant performance on different systems with various equipment and objectives, such as chiller plant \cite{vu2017data} and vacuum pumps \cite{jung2017vibration}, these successes may not be easily repeated in another setting, mostly due to the unaffordable cost to meet the expected data quality and quantity. Particularly, powerful machine learning models are usually hungry for big quality data, such as labels on equipment failure events, which involves huge human efforts in reading and annotating the historical data. To better address these demands on training data, we discuss two types of common limitations we meet in real-world applications.

\begin{figure}[t]
	\centering
	\includegraphics[width=\columnwidth]{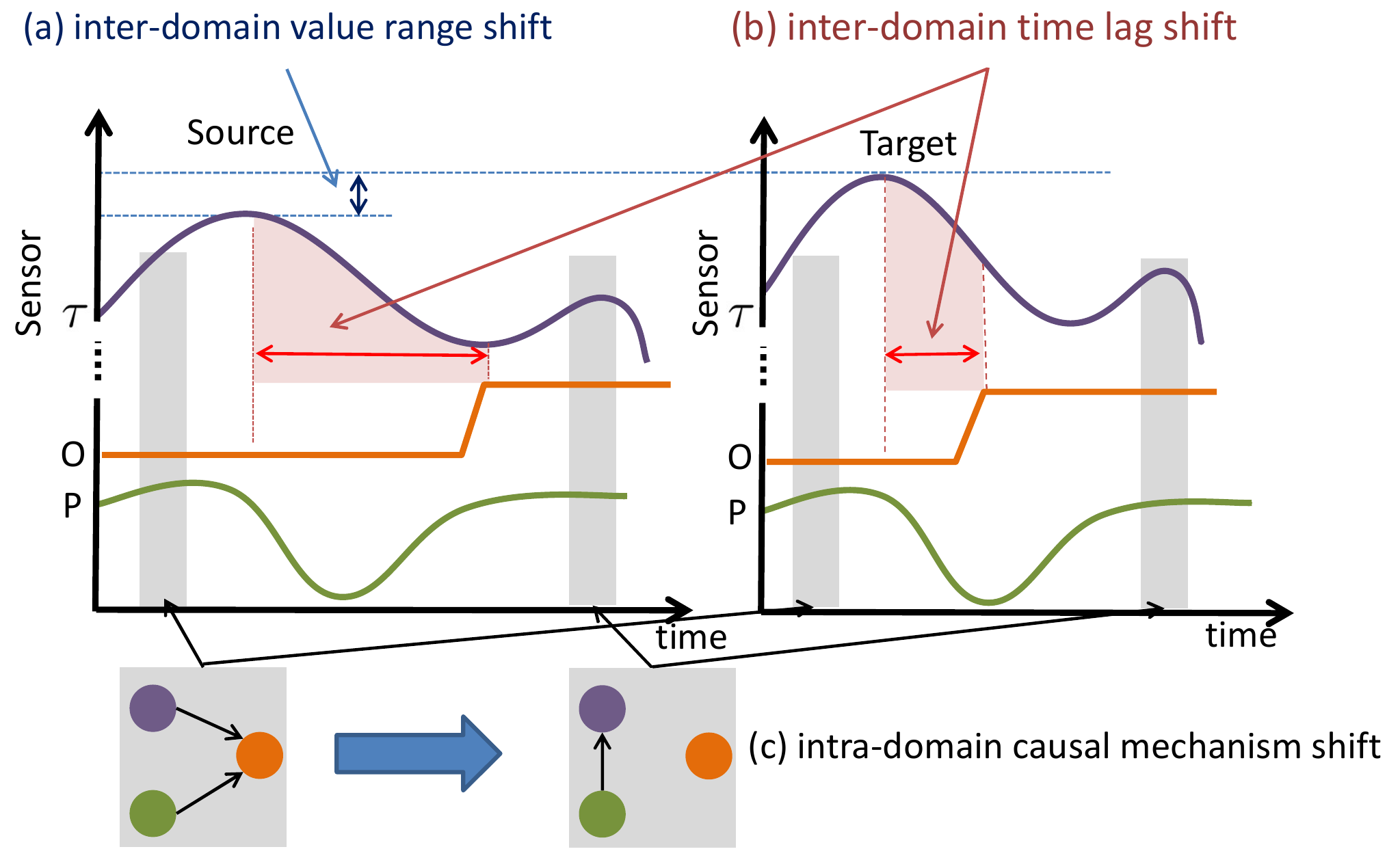}
	\caption{Illustration of three challenges of time series domain adaptation in boiler data. The purple, orange and green lines represent temperature ($\tau$) sensor values, operation status (O) sensor values and pressure (p) sensor values respectively. 
		(a) Inter-domain value range shift might appear in different domain, .e.g, the value range in source domain is smaller than target domain, which might leads to larger generalization error if we train with source domain data and test on target domain data directly. 
		(b) Inter-domain time lag shift caused by temporal causal mechanism appears in different domain, domain adaptation can capture the this causal mechanism and ignore the time lag simultaneously.
		(c) Intra-domain causal mechanism shift also exists in time series data. For example, the pressure and temperature jointly effect the operation status. This relationship between multivariate time series data should be captured by domain adaptation model.  (\textit{best view in color})}
	\label{fig:motivation}
\end{figure}

The first limitation is the lack of configuration coverage. In real-world mechanical systems, there are usually a variety of controlling parameters.
In chiller plant, for example, the parameters contain the variable speed drives(VSD) controlling, the frequencies of the pumps and cooling towers in the plant. Due to the limited variety of the conventional chiller plant control strategies, a chiller plant is operated under a small number of candidate configurations over the control parameters. This leads to a potential risk of overfitting of the data-driven model.
The second limitation is the lack of label coverage. In real-world IoT systems, events of interests, e.g., failures of pumps \cite{jung2017vibration}, are usually very rare. Every individual failure event, on the other hand, may cause huge financial loss. Supervised learning, however, builds reliable and meaningful models only when there are sufficient labelled data linked to the detection/prediction target event. 

Fortunately, domain adaptation, e.g., \cite{baktashmotlagh2013unsupervised,ganin2016domain,germain2016new,gong2016domain,Koniusz2016Domain}
which enables the system to reuse existing data from similar systems when building models over a new system with limited data by aligning the features and transforming the old model based on observations over these domains, can lift this restriction in our IoT applications.
While most of the existing approaches of domain adaptation are designed for non-sequential domain with fixed number of dimension, the neglect of temporal information is an important source of performance degradation, when these methods are applied on time series data directly. Recently, domain adaptation for time series data has received wide attention. \cite{purushotham2016variational} adversarially captures complex and domain-invariant temporal relationships by using variational recurrent neural network \cite{chung2015recurrent}. However, this method ignores the causal mechanism in time series data, because it mainly takes the hidden state of final time step into account instead of the hidden states of all the time steps.

In order to figure out the aforementioned challenges, we consider what can be transfered and what hinder the transferability in time series domain adaptation. Firstly, we assume that the causal mechanism is invariant. Because the physical mechanism is invariant among domains, and causal mechanism is a kind of this physical mechanism.
What's more, causal mechanism denotes the directed path between two random variables. In a word, a set of cause variables have impacts on the set of effect variables \cite{pearl2002causality}. According to our observation, there are two significant causal mechanisms of time series data in the mechanical systems.  One is dynamic causal mechanism, which means that one sensor value have influence on another sensor value in any time step.  
The other is temporal causal mechanism which means that the past values of one sensor value should contain information that helps predict another sensor value above. 

However, in order to transfer causal mechanism, three obstacles need to be tackled, which as shown in Figure \ref{fig:motivation}.
(1) Inter-domain value range shift means value ranges of sensors vary with domains. For example, the values range of the temperature sensors varies with the location of the machine. And the model that is trained by the machine with lower temperature range might not be suitable for the machine with higher temperature range.
(2) Inter-domain time lag shift means the time lags of causal effect vary with domains. According to the ideal gas law $PV=nR\tau$, where $P,V$ and $\tau$ are the pressure, volume and absolute temperature respectively. $R$ is the ideal gas constant and $n$ is the number of moles of gas. Because different boilers use different kinds of fuels, the ratio between temperature and pressure is different, which leads to different response time. 
(3) Intra-domain causal mechanism means that the causal mechanism between sensors in each time step. For example, in Figure \ref{fig:motivation}, the drop of temperature ($\tau$) causes the drop of of pressure ($P$) i.e. $\tau \rightarrow P$, and the operation status is decided by temperature and pressure jointly, i.e. $(\tau , P) \rightarrow O$.

In this paper, we utilize the invariant causal mechanism and solve the aforementioned obstacles by proposing a novel causal mechanism transfer network (\textbf{CMTN}). Firstly, because of different value ranges in different domains, we devise different feature extractors for the source and target domains separately. Then, we introduce two kinds of attention mechanisms to transfer two kinds of causal mechanisms according to our observations over the real data as shown in Figure \ref{fig:motivation}. In order to tackle the inter-domain time lag shift, we propose the transferable temporal attention mechanism. In order to tackle the intra-domain causal mechanism shift, we propose the transferable intra-sensors attention mechanism. Furthermore, 
We apply our \textbf{CMTN} to two case studies, including chiller plant optimization under lack of configuration coverage and boiler failure detection under lack of label coverage, and achieves significant improvements in modeling accuracy and consequently promising performance in their respective settings.

The rest of the paper is organized as follows. Section \ref{sec:related} 
reviews existing studies on time series modeling, domain adaptation, domain adaptation on time series as well as attention mechanism. 
Section \ref{sec:prelim} provides the problem definition on time series domain adaptation and adversarial domain adaptation model. 
Section \ref{sec:adaptation} proposes motivation based on the observation over the time series data in the mechanical systems and our causal mechanism transfer network for time series domain adaptation.
Section \ref{sec:cases} presents case studies on two completely different areas and conduct the ablation study on our CMTN. 
Section \ref{sec:conclu} concludes the paper with future work discussion. 

\section{Related Work}\label{sec:related}

In this section, we first review the existing techniques on time series modeling and domain adaptation, and then we give a brief introduction about time series domain adaptation and attention mechanism.

\noindent\textbf{Time Series: } Modeling and prediction on time series is a traditional research problem in computer science, with a number of successful cases, e.g., Autoregressive model 
\cite{forecastbook} 
and ARMA \cite{box1970distribution}
. With the introduction of domain expertise and graphical model, new approaches are proposed, e.g., 
\cite{qi2017mixture}, 
to enhance prediction accuracy.  The quick growth of computation power, on the other hand, has propelled the success of deep neural network models, e.g., RNN, LSTM \cite{hochreiter1997long}
and GRU \cite{chung2014empirical}, specifically designed for time series domain. In this paper, we adopt LSTM as our backbone network to model time series data.

\noindent\textbf{Domain Adaptation: }Unsupervised domain adaptation is a very important problem. The mainstream methods aim to extract the domain invariant feature between domains. Maximum Mean Discrepancy is one of the most popular methods by using kernel-reproducing Hilbert space \cite{borgwardt2006integrating,huang2007correcting,Ghifary2017Scatter}. Second-order statistics is proposed for unsupervised domain adaptation \cite{sun2016return}. Second or higher order scatters statistics can be used to measure alignment in CNN \cite{Koniusz2016Domain}.

Another essential approach in unsupervised domain adaptation is to extract the domain-invariant representation by introducing a domain adversarial layer for domain alignment. \cite{ganin2015unsupervised} introduces gradient reversal layer to fool the domain classifier and extracts the domain-invariant representation, \cite{tzeng2017adversarial} borrows the idea of generative adversarial network (GAN) \cite{goodfellow2014generative} and proposes a novel unified framework for adversarial domain adaptation.

Based on the adoption of causality view over the variables, the adaptation scenario can be determined by causal mechanism. \cite{Zhang2013Domain}
discusses three different application scenarios in domain adaptation. These scenarios respectively are target shift, condition shift and generalized target shift. Based on \cite{Zhang2013Domain}, \cite{Zhang2015Multi,gong2016domain} 
investigate more on the generalized target shift in the context of domain adaptation.

\noindent\textbf{Domain Adaptation on Time Series: }
Though unsupervised domain adaptation performs well in many tasks in computer version, there is limited work of domain adaptation in time series data. In NLP, \cite{seqlabelda} uses distributed representations for sequence labeling tasks. \cite{daclassification} simultaneously uses domain specific and invariant representations for domain adaptation in sentiment classification task while \cite{daembeddingclassification} solves the same problem by combining the generic embeddings with domain-specific ones. And \cite{purushotham2016variational} use variational method that produces a latent representation that captures underlying temporal latent dependencies of time series samples from different domains. However, this method extracts the domain-invariant representation with the final hidden state of RNN, which ignore the whole time series and its properties. In this paper, we proposed an unsupervised domain adaptation method for time series data, which extracts domain-invariant representation in the time-series level and consider the causal mechanism in time series data. What's more, we figure out time series domain adaptation in a causal view.

\noindent\textbf{Attention Mechanism: } Attention mechanism is also very significant in time series modeling. Motivated by how human beings pay visual attention to different regions of an image or correlate words in one sentence, attention mechanisms have become an integral part of network architectures in natural language processing and computer vision tasks. \cite{bahdanau2014neural} introduced a general attention mechanism into machine translation model which allow the model to automatically search for parts of the correlative words. \cite{li2017image} achieve promising performance in image caption by using a global-local attention method by integrating local representation at object level with global representation at image-level. Based on Transformer \cite{vaswani2017attention}, a general attention mechanism architecture, BERT \cite{devlin2018bert} achieves the state-of-the-art performance in question answering and language inference. Observing that not all region of an image is transferable, \cite{wang2019transferable} introduce attention mechanism into domain adaptation which focuses on transferable regions of an image.

In this paper, we introduce attention mechanism into time series domain adaptation, focusing on two kinds of transferable causal mechanism: dynamic causal mechanism and temporal causal mechanism. In this paper, we first present how the causal mechanisms happen in the time series by data observation, and then explain how to transfer this causal mechanism by introducing a dual attention mechanism.

\section{preliminary}\label{sec:prelim}

\subsection{Problem definition} 
We first denote $\bm{x}={\left[\bm{x}_1, \bm{x}_2, ...\bm{x}_t, ...\bm{x}_N\right]}$ as a multivariate time series sample with $N$ time steps, where $\bm{x}_t \in \mathbb{R}^M$ and $y \in \mathbb{R}$ as the certain label. When $y$ is a real number, the prediction on $y$ is a \emph{regression} problem over time series. When $y$ is a categorical value, it becomes a multi-class classification problem. 
We assume that $P^{S}\left(\bm{x},y\right)$ and $P^{T}\left(\bm{x}, y\right)$, which represent source domain and target domain respectively, have different distributions but share the same causal structure.  
$\left(\mathcal{X}^S, \mathcal{Y}^S\right)$ and $\left(\mathcal{X}^T, \mathcal{Y}^T\right)$ which are sampled from $P^S(\bm{x}, y)$ and $P^T(\bm{x}, y)$ separately, denote the source and target domain dataset. We further assume that each source domain time series sample $\bm{x}^S$ comes with $y^S$, while target domain has no labelled sample, and our goal is to devise a model that can predict label $y^T$ given time series sample $\bm{x}^T$ from target domain.

\subsection{Base Model}\label{sec:base}
We pick up \emph{recurrent neural network} model as the base approach for our time series modelling, because of its huge performance improvement over conventional approach \cite{chung2014empirical}. 
Specifically, we develop domain adaptation techniques based on Long Short-Term Memory (or LSTM in short) \cite{sak2014long}. In this subsection, we present the basic of LSTM and its usage in our target mechanical system. Formally, we define:
\begin{equation}
\begin{split}
 \left[\bm{h}_1, \bm{h}_2, \cdot \cdot \cdot, \bm{h}_N \right] = G_r\left(\bm{x};\theta_r\right), 
\end{split}
\end{equation}
in which $G_r$ denote the LSTM that accepts a time series sample as input and then outputs a time series hidden states and $\theta_r$ represent the parameters of LSTM. 

Dozens of domain adaptation algorithms, which are proposed in last decade, has shown significant performance improvement in their respective setting. We opt to use the strategy proposed by Ganin \cite{ganin2015unsupervised}. 
Generally speaking, their strategy models invariant features across domains by optimizing a domain predictor that is expected to fails to tell whether the extracted feature is from the source or the target domain. And we consider the feature extracted by aforementioned method is more robust for multiple domains.
One of the biggest benefits of the strategy is that the domain prediction loss, which denotes the loss of domain predictor, could be easily merged into the \emph{regression/classification} prediction loss, therefore enabling a holistic model training for both domain adaptation and label prediction optimization.

A straightforward solution to time series domain adaptation is to directly reuse existing algorithms originally designed for non-sequential data. Because the final hidden state $h_N$ is assumed to contain all the message of time series, so we take $h_N$ as the input of label predictor and domain predictor as shown in equation (2).
When training LSTM by using data from multiple domains, the objective loss function consists of two parts, the label loss for the source domain data and the domain prediction loss over both source and target domains. The label loss is used to minimize the error of LSTM when predicting the labels, while the domain prediction loss is used to control the alignment of features such that extracted features are consistent across domains.
\begin{equation}
\begin{split}
\hat{y_l} &= G_l\left(G_r\left(\bm{x};\theta_r\right);\phi_l\right) \\
\hat{y_d} &= G_d\left(G_r\left(\bm{x};\theta_r\right);\phi_d\right),
\end{split}
\end{equation}
in which $G_l$ represents label predictor with parameters $\phi_l$ and $G_d$ represents domain predictor with parameters $\phi_d$. The parameters $\phi_l$ and $\phi_d$ are trained by minimizing the following objective function. 
\begin{equation}
\begin{split}
&\mathcal{L}\left(\phi_l, \phi_d, \theta_r\right)\\
&=\frac{1}{n^S}\mathop{\sum_{x^S \in \mathcal{X^S}}}\mathcal{L}_y\left(G_l\left(G_r\left(\bm{x};\theta_r\right); \phi_l\right), y^S\right) \\
&-\frac{\lambda}{n^S+n^T}\mathop{\sum_{x \in \mathcal{X^S, X^T}}}\mathcal{L}_d\left(G_d\left(G_r\left(\bm{x};\theta_r\right); \phi_d\right), D\right).\\
\end{split}
\end{equation}

In which $D$ denotes the domain number, We let $D=0$ and $D=1$ as source and target domains labels, respectively.In next section, we will introduce our causal mechanism transfer network (CMTN) motivated by our data observation.

\section{model}\label{sec:adaptation}

The above base model only considers the alignment of the hidden representation of the data, while ignores the inherent properties of the time series data. Fortunately, we find that the causal mechanisms are invariant across the domains, due to the fact that all the machines from different domains still follow the same physical mechanism. Here the causal mechanism refers to a process that a cause contributes to the production of an effect. For example, as shown in Figure \ref{fig:motivation}, in the boiler system, the variation of temperature ($\tau$) causes the variation of pressure (P)(i.e. $\tau \rightarrow P$). Furthermore the temperature ($\tau$) and the pressure (P) effect the operation status jointly (i.e. $(\tau, P) \rightarrow O$ ).



Such invariant causal mechanism motivates our Causal Mechanism Transfer Network(\textbf{CMTN}) for time series domain adaptation-- extending the existing time series representation model with the casual mechanism of the data. Generally, we attempt to extend the sequence presentation model $G_r$ into two parts, the domain-invariant causal mechanism part and the domain-specific part. Formally, we extend $G_r\left(\bm x; \theta_r\right)$ to $G_r\left(\bm x; \theta_S, \theta_T,\theta_C\right)$, by splitting the parameters $\theta_r$ into three parts: $\theta_S$, $\theta_T$ and $\theta_C$. Among them $\theta_S$ and $\theta_T$ denotes the domain-specific parameters for the source and target domain respectively, and $\theta_C$ denote the domain-invariant parameters. 

However, it is still a challenging task to model the invariant causal mechanisms over the dynamic time series data, which is usually hindered by the following three phenomena of the limitations: inter-domain value range shift,  inter-domain time lag shift and intra-domain causal mechanism shift. These limitations come from our observation over the data. For example, as shown in Figure \ref{fig:motivation}, the value range of temperature of the chiller or boiler varies with the location of the machine; The time lag of causal effect (i.e. $\tau \rightarrow O$) varies with domains. 
The factors which effect the operation status can be more complex, for example temperature and pressure are jointly making effects on the operation status.
In the following, we will provide the details to solve the above three obstacles under the above general causal mechanism transfer framework.

\subsection{Domain Specific Feature Extractor}\label{sec:extractor}
\begin{figure}[!htbp]
	\centering
	\includegraphics[width=\columnwidth]{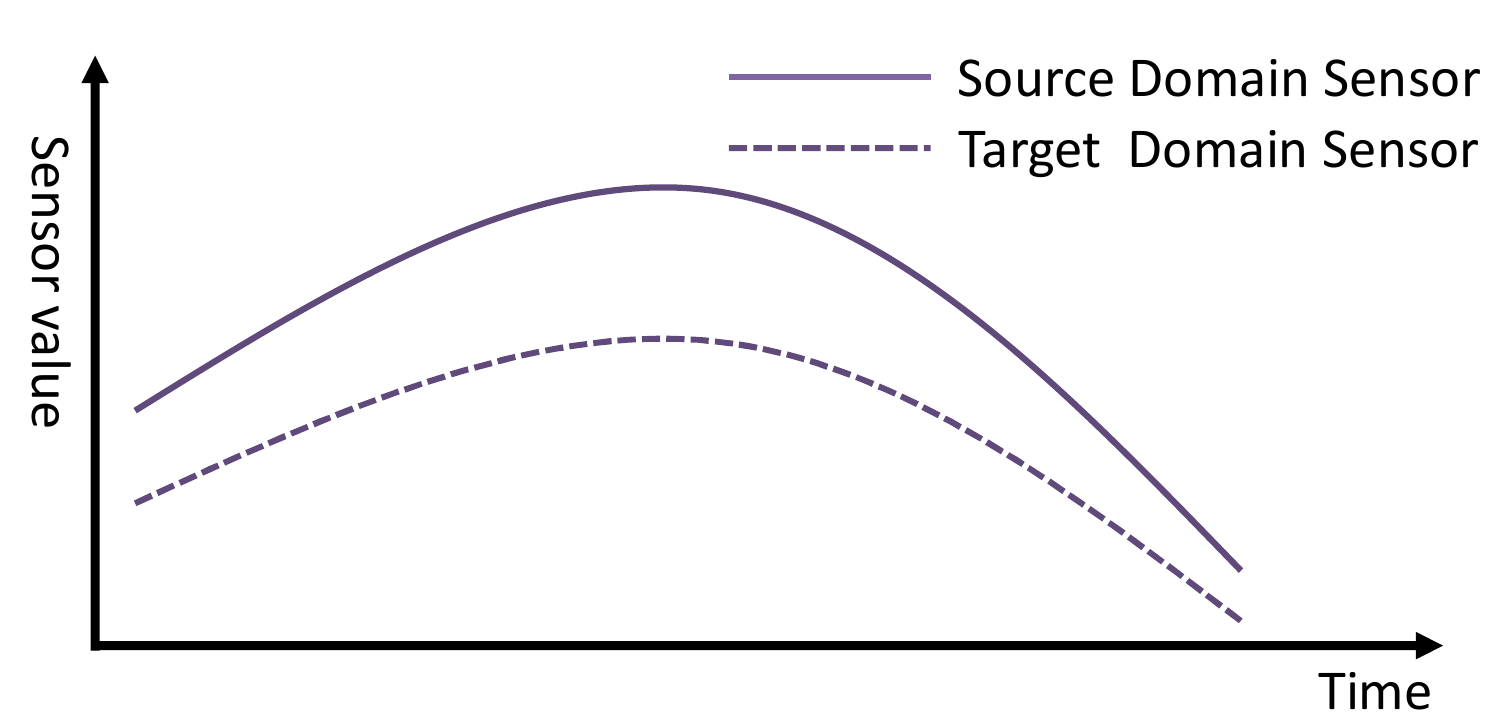}
	\caption{The illustration of value range shit. The value ranges of the source domains may be different from that of the target domains.}
	\label{fig:value_range}
\end{figure}

\noindent\textbf{Observation1 Inter-domain value range shift: }First of all, it is obvious that value range over the input vectors varies with different domain, which is shown in Figure \ref{fig:value_range}. In a boiler system, for example, the minimal and maximal values of certain sensor readings are very different from boiler to boiler. Traditional domain adaptation techniques, e.g. \cite{ganin2015unsupervised}, leave it to the feature alignment. It may affect the LSTM model which is shared by all domains when generating the features for final classification and regression task.

Motivated by our observations over varying value range over the input vectors in different domains, we insert a domain-specific feature extractor between the input $\bm{x}$ and LSTM. If we use Ganin's method \cite{ganin2016domain} directly, the shared LSTM that simply aligns the sensor readings of different value range will not achieve ideal performance. In our solution, we intentionally add a new layer for domain-specific feature extraction, i.e., the feature extractor in Figure \ref{fig:model}. It is expected to handle a wide spectrum of domain alignment problems by pre-processing the input values in an automatic manner. Formally, we have:
\begin{equation}
\begin{split}
\bm{f}^S=G^S_f\left(\bm{x}^S;\theta_S\right) \\
\bm{f}^T=G^T_f\left(\bm{x}^T;\theta_T\right)
\end{split}
\end{equation}
in which $G^S_f\left(\bm{x}^S;\theta_S\right)$ and $G^T_f\left(\bm{x}^T;\theta_T\right)$ are composed of a simple neural network respectively, $\theta_S, \theta_T$ are learnable projection matrices. $\bm{f}^S = \{\bm{f}^S_1, \bm{f}^S_2 \cdot \cdot \cdot \bm{f}^S_t \cdot \cdot \cdot  \bm{f}^S_N \}$ and $ \bm{f}^S_t \in \mathbb{R}^K$ is the feature generated by the source domain specific feature extractors. Similarly, we let $\bm{f}^T$ denote the feature generated by the target domain specific feature extractors, and further let $\bm{f}$ and $G_f\left(.\right)$ denote feature generated by any domain specific feature extractors and any domain specific feature extractors. Subsequently, we will take $\bm{f}$ as the input of the base model in section \ref{sec:base}.


As a summary, $\theta_S$ and $\theta_T$ in this section are domain-specific parameters, which are used to capture the different value range for the source and target domain respectively. $\{\theta_r\} \subset \theta_C$ are the domain-sharing parameters, which are used to model the domain-invariant causal mechanism.

\subsection{Transferable Temporal Causal Mechanism}
\begin{figure}[!htbp]
	\centering
	\includegraphics[width=\columnwidth]{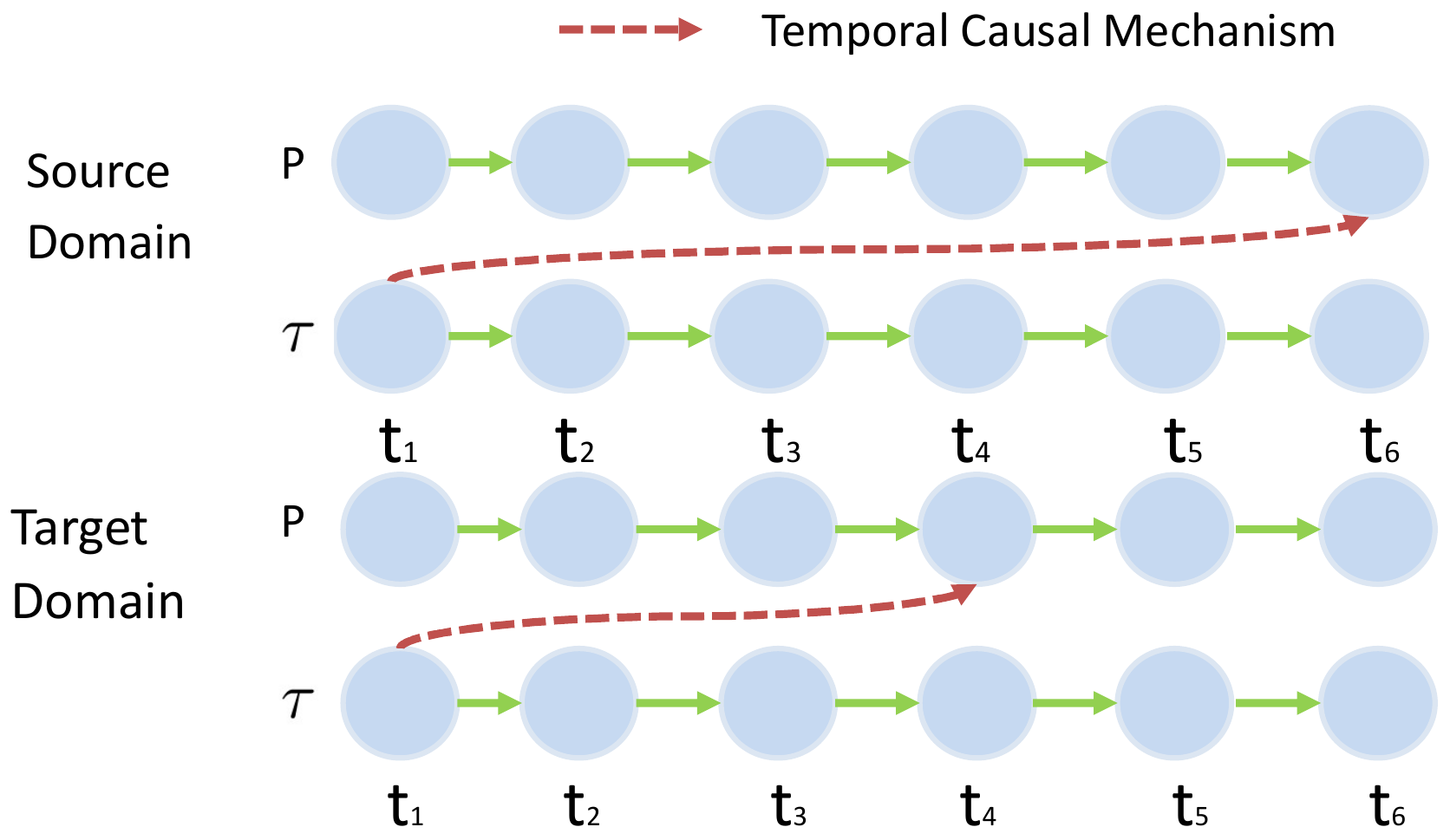}
	\caption{The illustration of temporal causal mechanism. The time lags vary across the domains, but the causal mechanism among the sensors (i.e., P is the cause of $\tau$ in the two domains) is transferable across the domains.}
	\label{fig:TCM}
\end{figure}

\noindent\textbf{Observation2 inter-domain time lag shift: } Temporal causal mechanism \cite{sun2008assessing,granger1969investigating,chikahara2018causal} is important to the modeling of multivariate time series data, for example, the relationship between temperature and pressure follows the Charles's law. However, because of the properties of different domain, such as the different degree of aging of different machines, there are time lags between different domains, which is shown in Figure \ref{fig:TCM}.

In mechanical system, the readings of sensors follow temporal causality, such as the relationship between the temperature and pressure. Formally, time series $A$ is said to be temporal-cause $B$ if it can be shown that those values of $A$ provide statistically significant information about future values of $B$. 

We can find that the ubiquity of temporal causality exists in the mechanism systems, but it comes with time lags due to properties of different domains. For example, in the chiller plant systems, the aging of pumps might lead to lags in response when the temperature is changing. In order to figure out this situation, we introduce the supervised attention mechanism that can select the relevant hidden states adaptively, i.e., by employing attention mechanism, the contributing hidden state might be assigned a larger weight, so the effectiveness of time lags will be negligible. Specifically, to calculate the context vector at time step $N$ over each hidden state before the final time step $N$, we define the weights of each hidden state $h_t$ as follow:
\begin{equation}
\begin{split}
u_t = \tanh & \left( \bm{h}_N W_e \bm{h}_t + b_e \right) \\
\gamma_t =& \frac{\exp\left( u_t \right)}{\sum_{t=1}^{N-1}\left( u_t \right)}\\
\widetilde{c} &= \sum_{t=1}^{N-1} \gamma_t h_t\\
c &= \left[\widetilde{c};\bm{h}_N\right],
\end{split}
\end{equation}
in which $W_e$ and $b_e$ are trainable parameters, and $\widetilde{c}$ is the candidate context vectors over all the hidden states except the last one. We generate the final context vectors by concatenating $\widetilde{c}$ and final hidden state $\bm{h}_N$. The aforementioned process is as follows:
\begin{equation}
\begin{split}
c = H_{TC}\left(\bm{h};W_e, b_e\right).
\end{split}
\end{equation}

As a summary, $\{ W_e, b_e\} \subset \theta_C$ are the domain-sharing parameters, which are used to model the transferable temporal causal mechanism proposed in this subsection.

\subsection{Transferable Dynamic Causal Mechanism}
\begin{figure}[!htbp]
	\centering
	\includegraphics[width=\columnwidth]{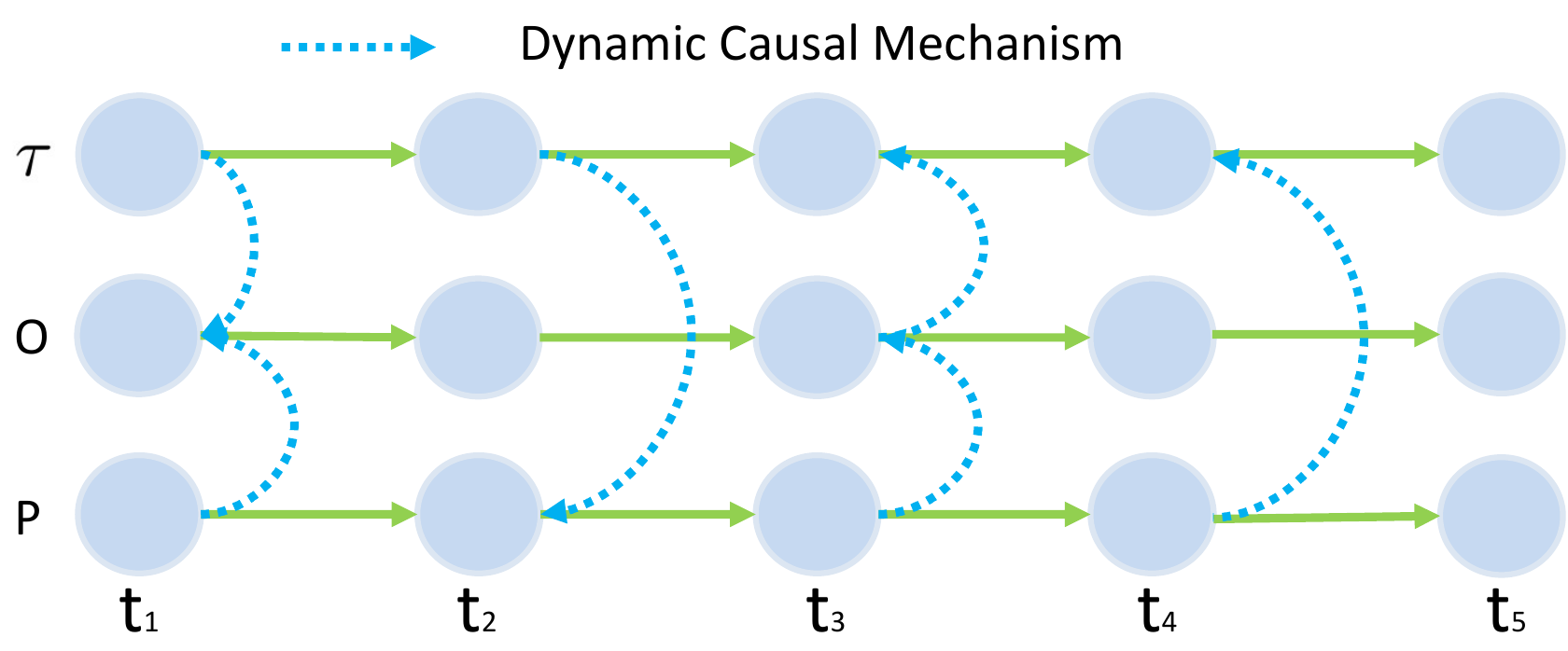}
	\caption{The illustration of dynamic causal mechanism. The causal mechanism among sensors change over time in a domain, but such mechanism is transferable across the domains.}
	\label{fig:DCM}
\end{figure}

\noindent\textbf{Observation3 intra-domain causal mechanism shift: }As shown in Figure \ref{fig:DCM}, we can find that the causal effect between sensors are changing over time, which depends on the sensor readings in the last time step. In chiller plant system, higher temperature leads to the increment of relative humidity, which further rev the chilled water pump, while lower temperature leads to the falloff of relative humidity, which further revs the condenser water pump. This causal effects are actually some physical mechanism, so it's reasonable to be transferred from the source domain to the target domain.

Next we introduce the transferable dynamic causal mechanism motivated by the aforementioned observation. In another word,  $\bm{f}^S$ and $\bm{f}^T$ share the same dynamic casual mechanism. To address this issue, given the $k$-th dimension of the $t$-th time step of domain specific extracted feature(i.e., $f_{t,k}$), we employ a self-attention mechanism that generates a transferable weight over sensors to adaptively capture the dynamic correlation of the multivariate time series data. Formally, we can calculate the weight of $k$-th feature at $t$-th time step (i.e., $\alpha_{t,k}$) by: 
\begin{equation}
\begin{split}
e_{t,k} =  v_d^\mathsf{T} \tanh & \left(W_d \left[ \bm{h}_{t-1};\bm{f}_t\right] + U_d \bm{f}_{t,k} + b_d \right) \\
\alpha_{t,k} &= \frac{\exp\left(e_{t,k}\right)}{\sum_{k=1}^{K}\left( e_{t,k} \right)} ,
\end{split}
\end{equation}
in which $v_d, W_d, b_d$ and $U_d$ are trainable parameters. The attention weights are jointly generated by the historical hidden state of LSTM $\bm{h}_{t-1}$ as well as current domain specific feature $\bm{f}_t$, and it also representation which sensor plays an important role in final prediction. Here, $\alpha_t$ as the vector of weights of each sensor. After generating the intra-sensors attention weight, the weighted sensor readings are calculated with:
\begin{equation}
\widetilde{\bm{f}_t} = \alpha_t^\mathsf{T} \bm{f}_t. 
\end{equation}

The aforementioned process is as follows:
\begin{equation}
\widetilde{\bm{f}} = H_{DC}\left(\bm{f};v_d, W_d, U_d, b_d\right).
\end{equation}

As a summary, $\{ v_d, W_d, U_d, b_d\} \subset \theta_C$ are the domain-sharing parameters, which are used to model the transferable dynamic causal mechanism proposed in this subsection. 

\subsection{Model Summary}
\begin{figure*}
	\centering
	\includegraphics[width=2.0\columnwidth]{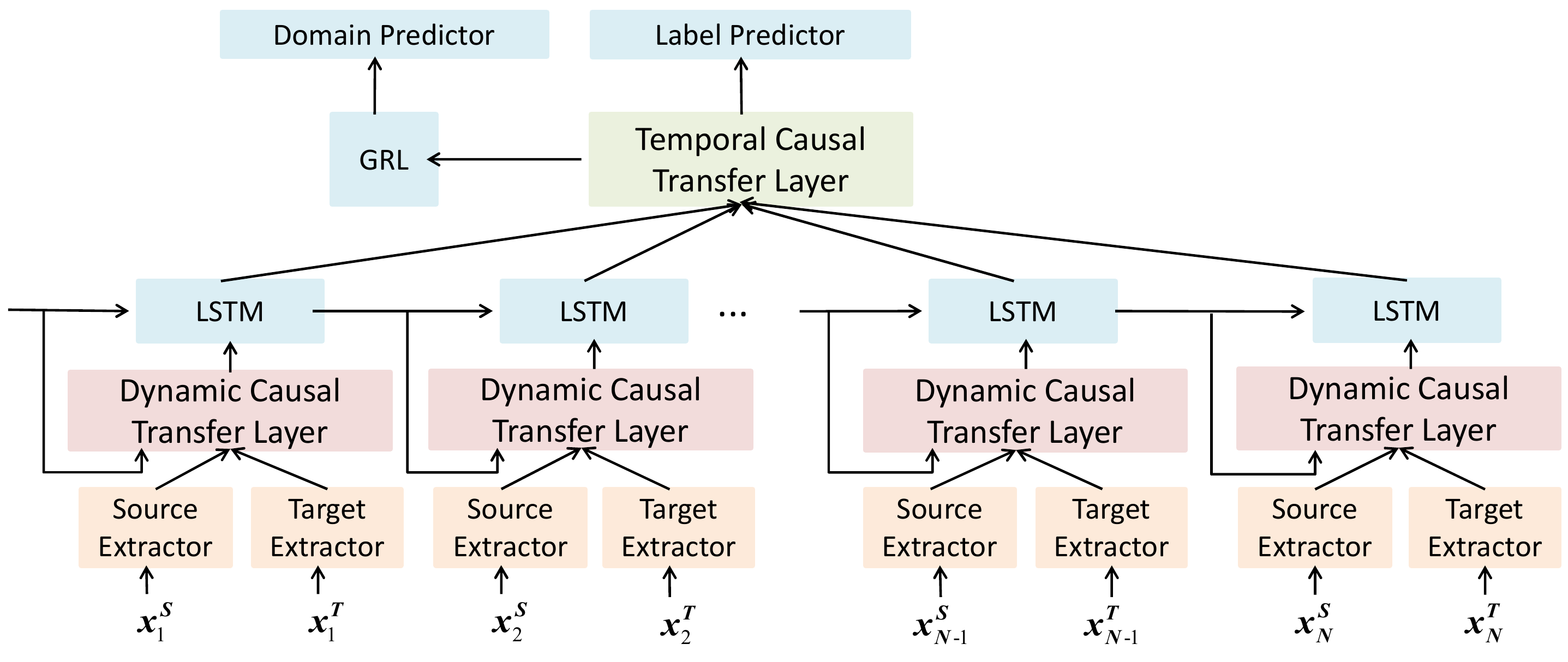}
	\caption{The architectures of Causal Mechanism Transfer Network (CMTN). From the input to the output, the domain specific feature extractor (in orange) employs a MLP layer to ease the mischief of inter-domain value range shift; the dynamic causal mechanism (in pink) employs a self-attention mechanism to capture the intra-domain causal mechanism shift; the temporal causal transfer layer employs a supervised attention layer to extract the important hidden state for the final prediction. (\textit{best view in color})}
	\label{fig:model}
\end{figure*}
The architecture of \textbf{CMTN} is shown in Figure \ref{fig:model}. 
First, We take the time series sensor value $\bm{x}$ as the input of domain-specific feature extractors, which mitigate the influence of different value ranges and the output of the extractors is feature $\bm{f}$. Second, the features are aligned by dynamic causal transfer layer which utilizes the feature $\bm{f}$ and the hidden state from the last time step $\bm{h}_{t-1}$ and we further get the weighted feature $\widetilde{\bm{f}}$. Third, by taking the hidden state from last time step $h_{t-1}$ and weighted feature $\widetilde{\bm{h}}$ as input, LSTM generates the hidden state $\bm{h}_t$. Fourth, by utilizing all the hidden states, the temporal causal transfer layer calculates the final context representation which not only contain all the message of the time series but also extract and highlight the most important state. Finally, we employ the gradient reversal layer to fool the domain predictor and the label predictor to generate the final decision.

The overall objective function of our approach is summarized as follows: 
\begin{equation}
\begin{split}
\small
&\mathcal{L}\left(\theta_C, \theta_S, \theta_T, \phi_d, \phi_l\right)\\
&\!=\!\frac{1}{n^S}\!\!\!\!\sum_{x^S \in \mathcal{X^S}}\!\!\!\mathcal{L}_y\!\left(\!G_l\!\left(\!H_{TC}\!\!\left(\!G_r\!\left(\!H_{DC}\!\left(\!G_{f}^S\!\!\left(\bm{x}^S\!;\!\theta_S,\!\theta_T\!\right)\!\right)\!\right)\!;\!\theta_C\!\right);\!\phi_l\right)\!, y^S\right) \\
&\!-\!\!\frac{\lambda}{n^S\!+\!n^T}\!\!\!\!\!\!\!\mathop{\sum_{x \in \mathcal{X^S, X^T}}}\!\!\!\!\!\!\mathcal{L}_d\!\left(\!G_d\!\left(\!H_{TC}\!\left(\!G_r\!\left(\!H_{DC}\!\left(\!G_f\!\!\left(\!\bm{x};\!\theta_S\!,\!\theta_T\!\right)\!\right)\!\right)\!;\!\theta_C\!\right)\!;\!\phi_d\!\right)\!,\! D\!\right)\!,\\
\end{split}
\end{equation}
where $n^S$ and $n^T$ is the size of source domain and target domain dataset, $\lambda$ is the parameter that trade-off the label prediction loss and the domain prediction loss in this unified optimization.

In the training procedure, we employ the stochastic gradient descent algorithm to find the optimal parameter set $\left(\hat{\theta_C}, \hat{\theta_S}, \hat{\theta_T}, \hat{\phi_d}, \hat{\phi_l}\right)$ as follows. In this procedure, all the samples are used, including the labelled source domain samples and the unlabelled target domain samples.
\begin{equation}
\begin{split}
\left(\hat{\theta_C}, \hat{\theta_S}, \hat{\theta_T}, \hat{\phi_d}, \hat{\phi_l}\right)
= \mathop{\arg\!\!\min_{\theta_C, \theta_S, \theta_T, \phi_d, \phi_l}}\!\!\!\!\mathcal{L}\left({\theta_C}, {\theta_S}, {\theta_T}, \phi_d, \phi_l \right).\\
\end{split}
\end{equation}

In the predicting procedure, we input the target domain samples into the model through the target feature extractor, and the labels of target domain samples are predicted as follows, 
\begin{equation}
y=G_l\left(H_{TC}\left(G_r\left(H_{DC}\left(G_{f}^T\left(\bm{x}^T;\hat{\theta_T}\right)\right) \right);\hat{\theta_C}\right); \hat{\phi_l}\right).
\end{equation}


\section{Case Studies and Experiment}\label{sec:cases}
In this section, the proposed CMTN method is experimental studied on two real-world applications: Chiller Plant Optimization and Boiler Fault Detection.

\subsection{Datasets}
\noindent\textbf{Chiller Plant Optimization:}
The chiller plant data which is provided by Kaer Pte. Ltd, consists of chiller plant sensor data collected from Building Management Systems (BMS) from two sites, each considered as one domain. The learning task is to predict total system power of a chiller plant, which is a regression problem, for energy optimization. We extract training data samples from the target domain, where the VSD speeds of condenser water pumps, chilled water pumps and fans of cooling towers are restricted to $95\%-100\%$ of allowed range. This is to simulate the situation at new chiller sites with insufficient data. Such data insufficiency is also common at chiller sites that have been running for years. We have encountered several chiller sites with VSD speeds set at a fixed speed for all the time.
The test data of the target domain contains data samples with full range of VSD speeds. Details of the dataset in terms of the start and end date, and the sizes of the source domain and the target domain are provided in Table~\ref{tb:data}. Table~\ref{tb:fchiller} lists all the features. Training and test data are split according to time. The first $80\%$ data are used as training data while the rest $20\%$ are used as test data.

Different from approaches in~\cite{vu2017data}
that decomposes a chiller plant into multiple components and models each component separately, we use a black-box approach based on LSTM to model the total system power. This is because it is less straightforward and even difficult to apply domain adaptation technique on a complex system with multiple inter-connected models.

\begin{table}
	\centering
	\caption{Duration and size of chiller data}\label{tb:data}
	\begin{tabular}{|l|c|c|c|}
		\hline
		& Start Date & End Date & Size \\ \hline \hline
		Source domain & 28/06/2017 & 14/07/2017 & 211K \\ \hline
		Target domain & 11/06/2017 & 14/07/2017 & 254K \\
		\hline
	\end{tabular}
\end{table}

\begin{table}
	\centering
	\caption{Features of chiller data}\label{tb:fchiller}
	\begin{tabular}{|l|l|}
		\hline
		Feature Name \\ \hline
		VSD speed of chilled water pump ($\%$) \\
		VSD speed of cooling tower fan ($\%$) \\
		VSD speed of condenser water pump ($\%$) \\
		Relative humidity ($\%$) \\
		Dry bulb temperature (outdoor) ($^\circ$C)\\
		System cooling load (RT)\\
		Number of chillers on\\
		Number of chilled water pumps on\\
		Number of cooling towers on\\
		Number of condenser water pumps on\\
		\hline
	\end{tabular}
\end{table}

\noindent\textbf{Boiler Fault Detection:}
The boiler data which is provided by SK Telecom, consists of sensor data from five boilers from 24/3/2014 to 30/11/2016. Each boiler is considered as one domain. The learning task is to predict \emph{faulty blow down valve} of each boiler. All the features used for this task is listed in Table~\ref{tb:fboiler}. In data pre-processing, we replace value $v_i$ with $v_i-v_{i-1}$ for columns with continuous increasing values along time, as indicated by ``delta'' in Table~\ref{tb:fboiler}. Notice that the boiler data is extremely unbalanced, as can be seen from the statistics of the five boilers listed in Table~\ref{tb:exp:boiler}. Less than $10\%$ of the total samples have faulty labels, with boiler 1 having $<2\%$ faulty samples. Due to lack of faulty labels, we use all the faulty data of source domains as training data for domain adaptation. To handle the extreme unbalance of the data, we apply down sampling on the normal samples of the source domain to obtain a balanced training dataset.
\begin{table}
	\centering
	\caption{Features of boiler data}\label{tb:fboiler}
	\begin{tabular}{|l|l|}
		\hline
		Feature Name \\ \hline
		Steam pressure main header \\
		Outdoor temperature \\
		Temperature concentrated water\\
		Operating time feed water (delta)\\
		Temperature exhaust gas\\
		Volume feed water (delta)\\
		Temperature feed water\\
		Temperature tube wall\\
		Damper angle\\
		Temperature scale\\
		Temperature external\\
		Operating status\\
		Operating code\\
		Input status\\
		Power usage meter (delta)\\
		Steam pressure\\
		Operating time chemical injection (delta)\\
		Combustion time (delta)\\
		Number of ignition (delta)\\
		Gas consumption\\
		\hline
	\end{tabular}
\end{table}

\begin{table}
	\centering
	\caption{Statistics of the boiler data. `Ratio' is the ratio of \# normal samples over \# of samples.}\label{tb:exp:boiler}
	\begin{tabular}{|c|c|c|c|}
		\hline
		Boiler ID & \# of samples & \# of faulty samples & Ratio \\ \hline \hline
		1 & 89969 & 1334 & 0.98 \\
		2 & 90120 & 7170 & 0.92 \\
		3 & 83145 & 1168 & 0.98 \\
		4 & 89718 & 4936 & 0.94 \\
		5 & 89639 & 6712 & 0.92 \\
		\hline
	\end{tabular}
\end{table}
\subsection{Evaluaion Metrics}
We use application specific criteria to evaluate the performance of our model and the baselines. For Chiller Plant Optimization case, we use the mean absolute percentage error (MAPE) to evaluate the performance of proposed model. MAPE is formally defined as follows:
\begin{equation}
\text{MAPE}=\frac{100}{n}\sum_{i}^{n}\left|\frac{y_i-\hat{y_i}}{y_i} \right|,
\end{equation}
where $y_i$ is the actual value and $\hat{y_i}$.
For Boiler Fault Detection, we use another two criteria to evaluate the performance of boiler fault detection:
\begin{itemize}
	\item Accuracy of fault detection as the percentage of correctly predicted samples.
	\item Area under the curve (AUC) of the correctly predicted faulty samples.
\end{itemize}
It is worth noting that we report the AUC over the fault samples in our experiment. As the boiler data is extremely unbalanced, a prediction model that always predicts 'normal' could achieve $>90\%$ accuracy and AUC over the fault samples could enable us to have a better understanding of the performance of the model.

\subsection{Baselines}
We compare our approach against the following baselines:
\begin{itemize}
	\item \textbf{LSTM\_S2T} uses source domain data to train a LSTM model and apply it on the target domain without any adaptation(S2T stands for source to target).It is expected to provide the lower bound performance.
	\item \textbf{Ganin} implements the domain adaptation architecture proposed in \cite{ganin2015unsupervised} with GRL(Gradient Reversal Layer) on LSTM, which is a straightforward solution for time series domain adaptation.
	\item \textbf{VRADA} implements the domain adaptation architecture proposed in \cite{purushotham2016variational} which combines the GRL with VRNN \cite{chung2015recurrent}. However, it only aligns the the final latent representation from recurrent latent variables model. 
\end{itemize}
Besides the above baselines, we also consider three variations of our approach to evaluate the effect of individual component as:
\begin{itemize}
	\item \textbf{CMTN-NDE}: We only remove the domain specific extractors.
	\item \textbf{CMTN-NGA}: We only remove the temporal causal transfer layer.
	\item \textbf{CMTN-NLA}: We only remove the dynamic causal transfer layer.
\end{itemize}
Our model and the baselines are implemented with Tensorflow \cite{girija2016tensorflow} on the server with one GTX-1080 and Intel 7700K. We set the length of time series sample as 6, i.e. $N=6$. The setting of each model are provided in Table \ref{tb:exp:parachiller}.
\begin{table}
	\centering
	\caption{Settings of Models on Chiller Data}\label{tb:exp:parachiller}
	\begin{tabular}{|p{3cm}|c|p{0.6cm}|p{0.9cm}|p{0.6cm}|}
		\hline
		& LSTM\_S2T & Ganin &VRADA& CMTN \\ \hline \hline
		Batch size & 512 & 512 & 512& 512 \\
		LSTM hidden layer size & 500 & 500 & 500& 500 \\
		LSTM layer & 1 & 1 & 1& 1 \\
		MLP hidden layer size & 100 & 100 & 100& 100 \\
		MLP layer & 1 & 1 & 1& 1 \\
		Domain specific feature size & 100 & 100 & 100& 100 \\
		Optimizer & Adam & Adam & Adam& Adam \\
		Learning rate & 0.0001 & 0.003 & 0.003& 0.003 \\
		Coefficient & - & 0.0001 & 0.0001& 0.005 \\
		Dropout rate & 0.5 & 0.1 & 0.2& 0.1 \\
		\hline
	\end{tabular}
\end{table}

\subsection{Results on Chiller Plant Optimization}
\begin{table}
	\centering
	\caption{MAPE on total system power prediction on chiller data}\label{tb:exp:chiller}
	\begin{tabular}{|l|c|}
		\hline
		Method & MAPE (\%) \\ \hline \hline
		LSTM\_S2T & 371.86 \\
		Ganin & 4.71 \\
		VRADA & 4.21 \\
		CMTN-NDE & 3.97 \\
		CMTN-NGA & 3.34 \\
		CMTN-NLA & 3.41 \\
		CMTN & \textbf{3.28}\\
		\hline
	\end{tabular}
\end{table}
\textbf{Accuracy of the system power prediction:} The MAPE of all model for total system power prediction are reported in Table \ref{tb:exp:chiller}. 
Our approach achieve the lowest MAPE among all models. It's $30.4\%$ lower that of Ganin and $22.1\%$ lower than that of VRADA. The MAPE of CMTN-NDE is $15.7\%$ and $5.7\%$ lower than that of Ganin and VRADA respectively. This indicates the effectiveness of transferable temporal and dynamic causal mechanism, which is different from that in Ganin and VRADA. The MAPE of LSTM\_S2T is the worst, which simply implies that applying source domain knowledge directly to target domain without adaptation is not going to work on the chiller plant. 

\textbf{Power saving after using the power prediction:} In order to evaluate the usefulness of domain adaptation models on energy saving, we conduct simulation of real-time VSD speed optimization on the test data of target domain as proposed in \cite{vu2017data}. The main idea is to search for optimal VSD speeds of pump and fans every $k$ time steps with the minimum total system power based on the domain adaptation models, assuming other features (e.g., weather, cooling load, etc) remain the same.

\begin{figure}[!htbp]
	\centering
	\includegraphics[width=\columnwidth]{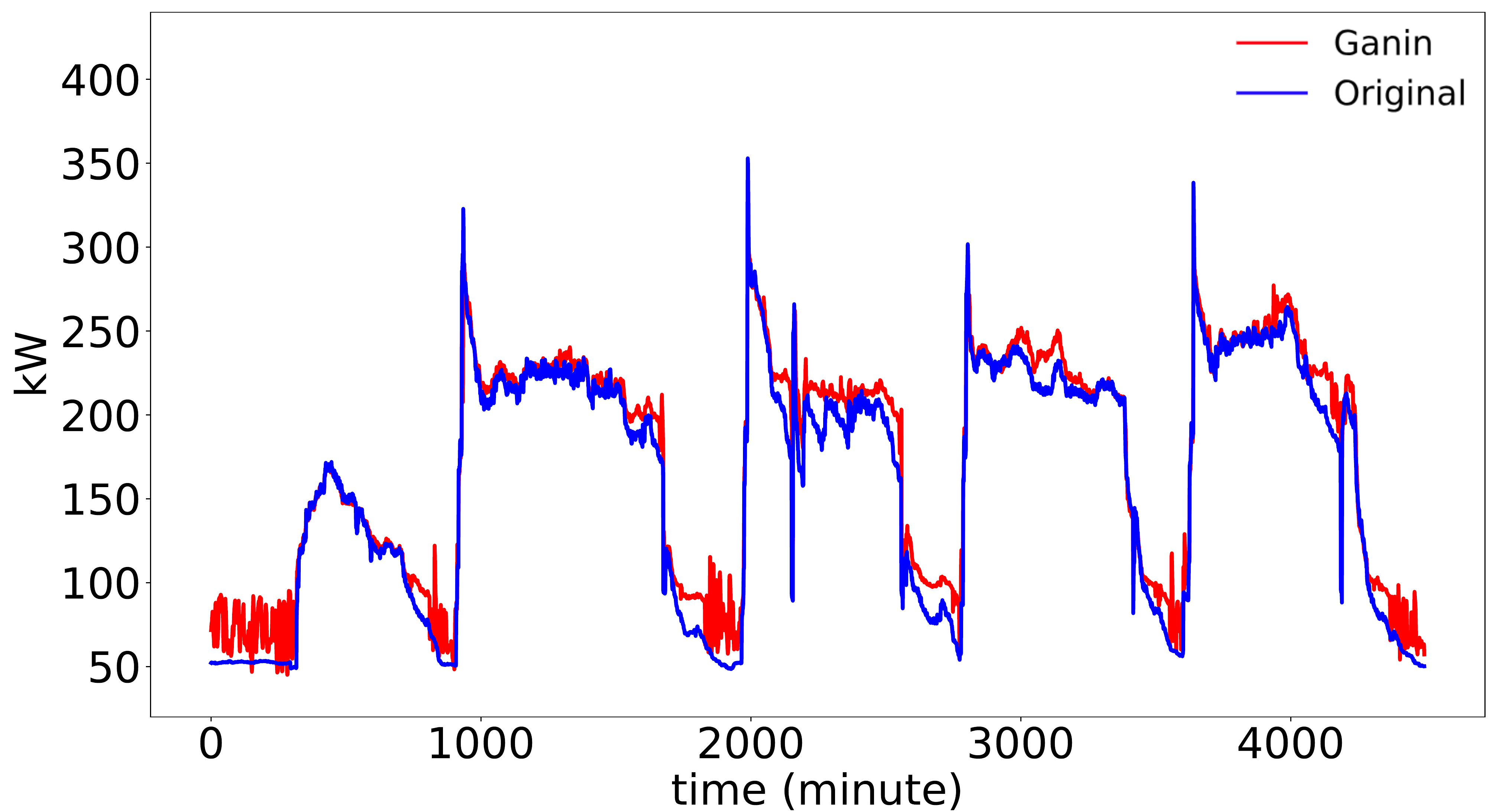}
	\caption{Comparison of original total system power with that of Ganin’s approach with optimization.}
	\label{fig:ganinsaving}
\end{figure}

\begin{figure}[!htbp]
	\centering
	\includegraphics[width=\columnwidth]{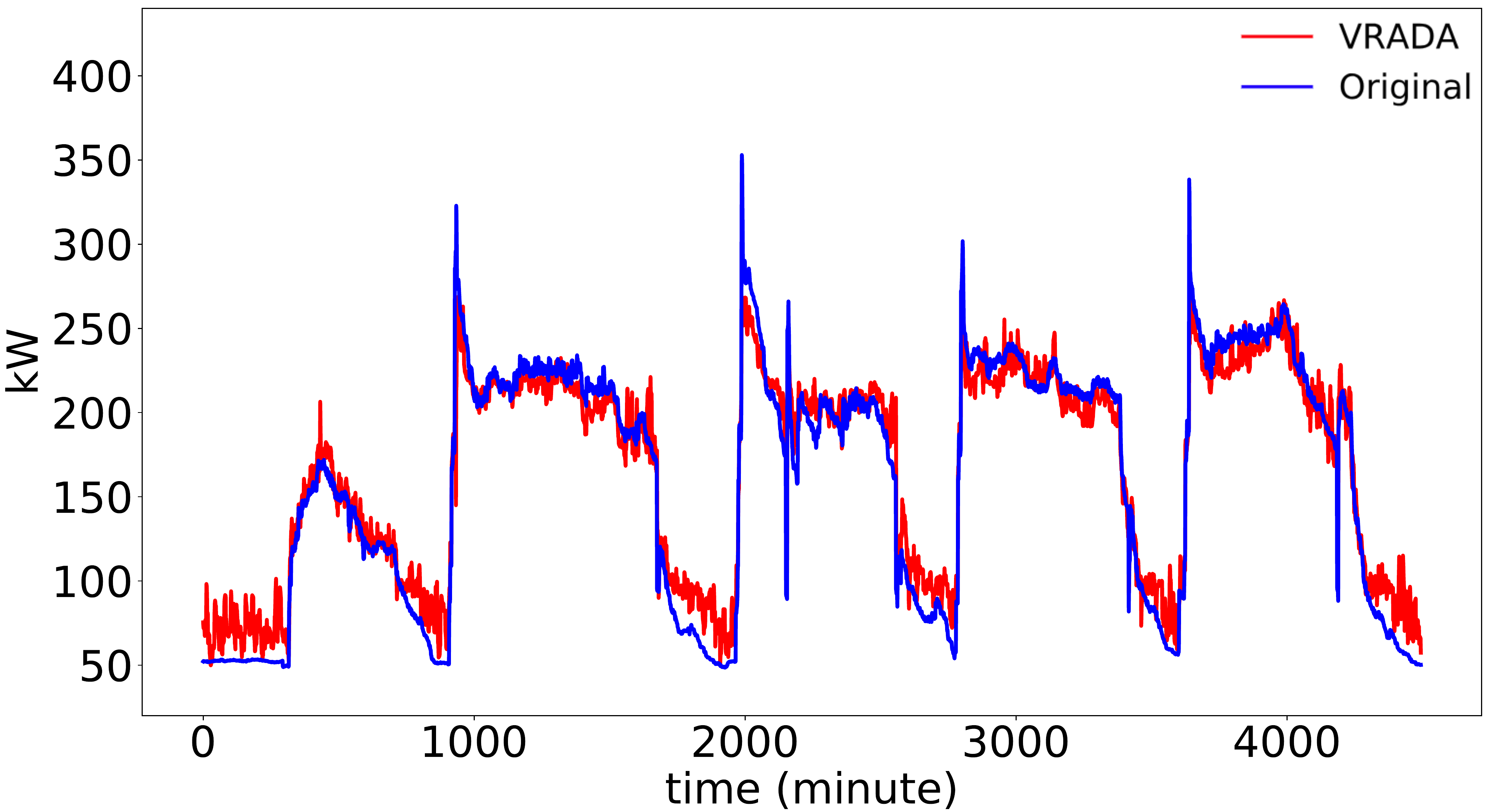}
	\caption{Comparison of original total system power with that of VRADA’s approach with optimization.}
	\label{fig:VRADA}
\end{figure}

\begin{figure}[!htbp]
	\centering
	\includegraphics[width=\columnwidth]{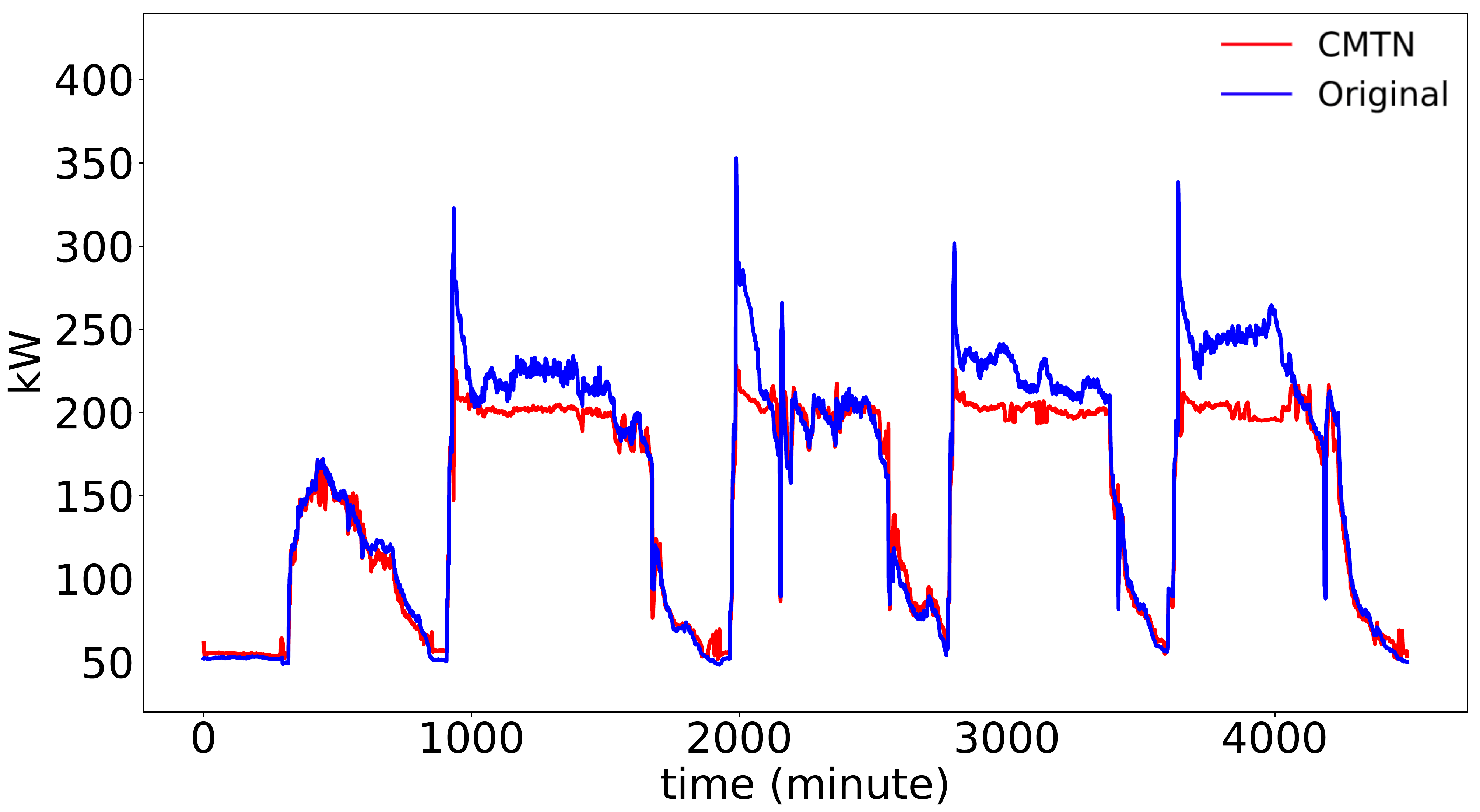}
	\caption{Comparison of original total system power with that of CMTN’s approach with optimization.}
	\label{fig:oursaving}
\end{figure}

\begin{table}
	\centering
	\caption{Percentage of total system power saving on chiller data}\label{tb:exp:optchiller}
	\begin{tabular}{|l|c|c|}
		\hline
		Model & Energy (kWh) & Energy Difference (\%) \\ \hline \hline
		Original & 15858 &   \\
		\hline
		Ganin & 17385 & +9.62 \\
		VRADA & 16971 & +7.02 \\
		CMTN-NDE & 16930 & +6.76 \\
		CMTN-NGA & 16003 & +0.91 \\
		CMTN-NLA & 16535 & +4.28 \\
		CMTN & \textbf{15532} & \textbf{-2.05} \\
		\hline
	\end{tabular}
\end{table}
Upon finding the optimal speed, we first train a LSTM\_T2T model, which is trained and tested with target domain training and test dataset respectively. And then we apply the most accurate LSTM\_T2T model to predict the corresponding total system power and compare it against the original power. The result of Ganin, VRADA and our apporach are plotted in Figure \ref{fig:ganinsaving}, \ref{fig:VRADA} and \ref{fig:oursaving} respectively, with 5-day simulations covering4 weekdays and 1 weekend day. Note that the energy consumption of original setting is already optimization outcomes of our previous data-driven method in \cite{vu2017data}.

Our approach with optimization is able to further reduce energy consumption, by consistently reaching lower power in most of the cases as show in Figure \ref{fig:oursaving}, while Ganin's approach generates similar or even higher power after optimization due to it's MAPE on power prediction.

The corresponding energy consumption (kWh) and percentage of energy saving in total system power, if possible, of all models are reported in Table \ref{tb:exp:optchiller}. Due to the high requirement on accurate modeling, only our approach is able to achieve energy saving by $2.05\%$ in the simulation. With electricity tariff being around SGD\$0.20, the optimization based on our domain adaptation model can save roughly SGD\$65.2 in five days. Since all domain adaptation techniques tested here do not use any labels from target domain, the saving achieved by our approach is significant.

\subsection{Results on Boiler Fault Detection}
\textbf{Accuracy of the boiler fault detection:} We use boiler 4 as the source domain, which has the median number of fault labels among the five boilers. The rest of the boilers are used as target domains. We report AUC of each source-target pair in Tables \ref{tb:boiler:4-1}, \ref{tb:boiler:4-2}, \ref{tb:boiler:4-3} and \ref{tb:boiler:4-5} respectively.

\begin{figure}[!htbp]
	\centering
	\includegraphics[width=\columnwidth]{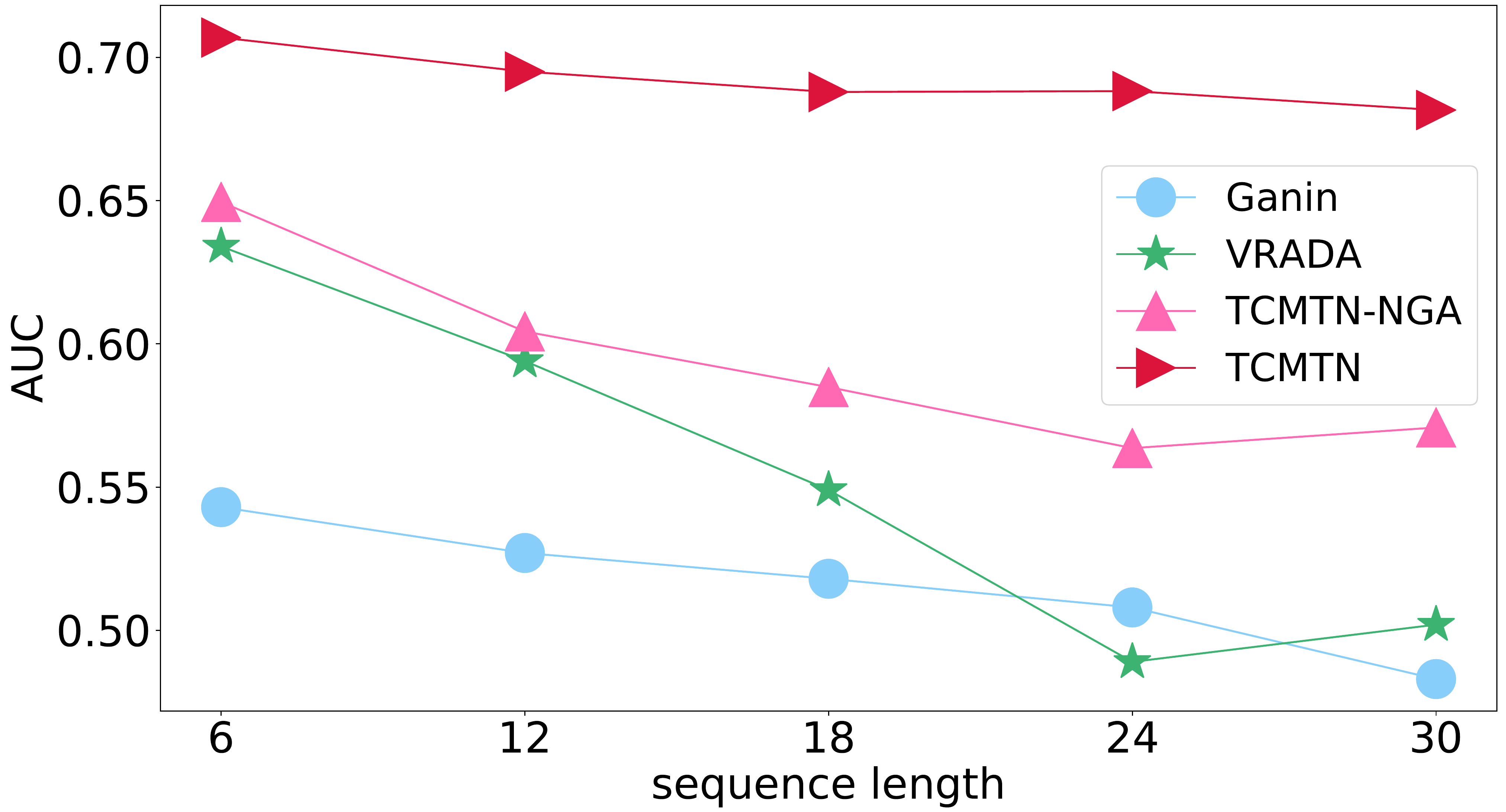}
	\caption{Comparison of AUC with different length of time series input under the setting $4 \rightarrow 1$.}
	\label{fig:global_attn_seq_len}
\end{figure}

\begin{table}
	\centering
	\caption{Results on Boiler 4 (Source) and 1 (Target)}\label{tb:boiler:4-1}
	\begin{tabular}{|l|c|c|}
		\hline
		Method & Accuracy & AUC \\ \hline \hline
		LSTM\_S2T & 0.970 & 0.475 \\
		Ganin & 0.975 & 0.533 \\
		VRADA & 0.985 & 0.634 \\
		CMTN-NDE & 0.982 & 0.640 \\
		CMTN-NGA & 0.980 & 0.642 \\
		CMTN-NLA& 0.985 & 0.6763 \\
		CMTN & \textbf{0.985} & \textbf{0.707} \\
		\hline
	\end{tabular}
\end{table}

\begin{table}
	\centering
	\caption{Results on Boiler 4 (Source) and 2 (Target)}\label{tb:boiler:4-2}
	\begin{tabular}{|l|c|c|}
		\hline
		Method & Accuracy & AUC \\ \hline \hline
		LSTM\_S2T & 0.940 & 0.864 \\
		Ganin & 0.971 & 0.909 \\
		VRADA & 0.972 & 0.934 \\
		CMTN-NDE & 0.976 & 0.926 \\
		CMTN-NGA & 0.975 & 0.925 \\
		CMTN-NLA& 0.971 & 0.947 \\
		CMTN & \textbf{0.977} & \textbf{0.948} \\
		\hline
	\end{tabular}
\end{table}

\begin{table}
	\centering
	\caption{Results on Boiler 4 (Source) and 3 (Target)}\label{tb:boiler:4-3}
	\begin{tabular}{|l|c|c|}
		\hline
		Method & Accuracy & AUC \\ \hline \hline
		LSTM\_S2T & 0.978 & 0.300 \\
		Ganin & 0.979 & 0.475 \\
		VRADA & 0.986 & 0.720 \\
		CMTN-NDE & 0.982 & 0.534 \\
		CMTN-NGA & 0.978 & 0.709 \\
		CMTN-NLA& 0.986 & 0.800 \\
		CMTN & \textbf{0.986} & \textbf{0.877} \\
		\hline
	\end{tabular}
\end{table}

\begin{table}
	\centering
	\caption{Results on Boiler 4 (Source) and 5 (Target)}\label{tb:boiler:4-5}
	\begin{tabular}{|l|c|c|}
		\hline
		Method & Accuracy & AUC \\ \hline \hline
		LSTM\_S2T & 0.975 & 0.930 \\
		Ganin & 0.967 & 0.932 \\
		VRADA & 0.980 & 0.945 \\
		CMTN-NDE & 0.969 & 0.936 \\
		CMTN-NGA & 0.976 & 0.929 \\
		CMTN-NLA& 0.981 & 0.949 \\
		CMTN & \textbf{0.986} & \textbf{0.954} \\
		\hline
	\end{tabular}
\end{table}

Overall, our approach achieves the highest accuracy and AUC on all setting. It outperforms Ganin and VRADA by improving the AUC over faulty samples, for example, by $84.6\%$ for Ganin(from 0.475 to 0.877 in Table \ref{tb:boiler:4-3}) and by $21.8\%$ for VRADA(from 0.720 to 0.877 in Table \ref{tb:boiler:4-3}) on pair Boiler 4 and Boiler 3 (denoted by $4 \rightarrow 3$). All models perform well on pair $4 \rightarrow 5$ and $4 \rightarrow 2$. Even LSTM\_S2T can achieve AUC over 0.930 and 0.864 respectively. This is probably because these boilers (i,e., boiler 2, 4 and 5) encounter similar problems, i.e., faulty blow down valve, after installation. Therefore they tend to share more common properties without adaptation that may result in fault due to issues in installation. Even in such case, domain adaptation is able to further improve the accuracy and AUC, for example, by $9.72\%$ for pair $4 \rightarrow 2$ than LSTM\_S2T.

However, the performance on pair $4 \rightarrow 3$ and $4 \rightarrow 1$ are much worse than the other cases. The highest AUC over faulty samples on pair $4 \rightarrow 1$ is only 0.707(Table \ref{tb:boiler:4-1}). The reasons are two fold: first, these two target domains, i.e., Boiler 1 and Boiler 3, contain much fewer faulty labels than the others. This makes it more difficult to learn domain specific feature extractor. Second, these two boilers do not encounter 'faulty blow down valve' problems after installation. Thus they tend to share less similar properties with the source domain.

However, the improvement over AUC of LSTM\_S2T by our domain adaptation approach is significant in such case, e.g., by $192.33\%$ on $4 \rightarrow 3$ and $48.64\%$ on $4 \rightarrow 1$, though they have not yet reached the level for reliable industrial adoption. Inspired by these observations, a possible solution for quick examination of whether domain adaptation technique would apply on a new domain is to use S2T as the baseline. If S2T can achieve reasonable performance, it shows higher chances to obtain a promising result with domain adaptation. We leave this as our feature work.

\subsection{Ablation Study} 

\textbf{Study on the domain specific feature extractor:} The value ranges of some sensors of each boiler with wide difference are shown in Table \ref{tab:performance_comparison}, and we can find that  boiler 3 contains the largest otherness of the value range among all the boilers. At the same time, the experimental result reveal that CMTN-NDE, which removes the domain specific extractors, gains significant drop over baselines compared with CMTN and even gets a lower AUC score than VRADA. From the result of boiler fault detection, we observe that: 1) Different value range of sensors can lead to negative transfer. 2) The domain specific feature extractors can mitigation the domain-variant influence.
\begin{table}
	
	\centering
	
	\caption{some sensor value ranges with large otherness}
	\label{tab:performance_comparison}
	\resizebox{0.45\textwidth}{13mm}{
		\begin{tabular}{|p{0.6cm}|p{1.9cm}|p{1.6cm}|p{1.4cm}|p{1.5cm}|}
			\hline
			Boiler&Operating time feed water&Temperature Exhaust Gas&Power usage meter&Temperature Tube Wall\cr
			\hline
			\hline
			1&$0 \sim 4$&$0 \sim 122$&$0 \sim 92.76$&$20 \sim 172$\cr\hline
			2&$0 \sim 4$&$0 \sim 126$&$0 \sim 74.43$&$21 \sim 176$\cr\hline
			3&$0 \sim \textbf{9}$&$0 \sim \textbf{413}$&$0 \sim \textbf{149.81}$&$19 \sim \textbf{199}$\cr\hline
			4&$0 \sim 3$&$18 \sim 132$&$0 \sim 62.03$&$21 \sim 175$\cr\hline
			5&$0 \sim 1$&$19 \sim 127$&$0 \sim 43.81$&$19 \sim 183$\cr\hline
		\end{tabular}}
\end{table}
	
\textbf{Study on the transferable temporal causal mechanism:} Motivated by the fact that temporal causal mechanism keeps invariable among domains while time lag varies, we adopt attention mechanism for transferable temporal causal mechanism module, which not only consider the final hidden state, but also the others. Longer the input time series is, less information about preceding information is included in the final hidden state. Therefore, we evaluate the effect transferable temporal causal mechanism module by taking time series with different length as input, the experiment is shown in Figure \ref{fig:global_attn_seq_len}.
	
According to the result, we can observe that: 1)The performance of TCMTN-NGA is still better than VRADA and the longer the sequence length, the larger the gap between TCMTN-NGA and VRADA, which reflect the useless of domain specific extractors and transferable dynamic causal mechanism. 
2)The AUC of Ganin, VRADA and TCMTN-NGA drop sharply with the increasement of the length of the time series while slope of CMTN is much small than other compared approach. This is because CMTN applies temporal causal mechanism to all the hidden state, which utilizes all the hidden states and decreases the effect of domain-variant time lag and capture the temporal causality between time series at the same time. Though VRADA can capture complex and domain-invariant temporal relationships, it fails in time-series level feature alignment, so the increasement of sequence length will make a great impact on transferability.
	
\textbf{Study on the transferable dynamic causal mechanism: } As shown in Table \ref{tb:boiler:4-1}, \ref{tb:boiler:4-2}, \ref{tb:boiler:4-3} and \ref{tb:boiler:4-5}, we observe that: 1) the combination of domain specific extractors and transferable temporal causal mechanism shows superiority against VRADA, especially in $4 \rightarrow 3$. 2) After appending the dynamic temporal causal mechanism, the experiment result improves ulteriorly, which demonstrates the importance of transferable dynamic causal mechanism. VRADA and Ganin simple consider that the weight of each sensor in each time step are the same, and the main drawback is that some sensor value might be useless and even have interference effect to detection. 
	
	
\section{Conclusion}\label{sec:conclu}
In this paper, we present novel Casual Mechanism Transfer Network for time series domain adaptation. We demonstrate the usefulness of the approach on two real-world case studies on mechanical systems. The case studies show positive results on model performance improvement even when the mechanical system lacks labels over historical data. By deploying these data-driven models, we are capable of reducing energy consumption of chiller plant and accurate detection of boiler failures. 
Furthermore, we not only mitigate the different value ranges and time lags among different machines in mechanism system, but also exploit the causal mechanisms among time series data to transfer the knowledge from source domain to target domain.
	

\ifCLASSOPTIONcaptionsoff
  \newpage
\fi



\bibliographystyle{IEEEtran}
\bibliography{IEEEabrv,chiller}


%








\end{document}